\documentclass{article}

\usepackage{arxiv}

\usepackage[utf8]{inputenc} 
\usepackage[T1]{fontenc}    
\usepackage{hyperref}       
\usepackage{url}            
\usepackage{booktabs}       
\usepackage{amsfonts}       
\usepackage{nicefrac}       
\usepackage{microtype}      
\usepackage{lipsum}		
\usepackage{graphicx}
\usepackage{natbib}
\usepackage{doi}

\usepackage{times}  
\usepackage{helvet}  
\usepackage{courier}  
\usepackage{caption} 
\usepackage{algorithm}
\usepackage{algorithmic}
\usepackage{newfloat}
\usepackage{listings}

\usepackage{multirow}
\usepackage{cuted}     
\usepackage{lipsum}     
\usepackage{cleveref}
\usepackage[table]{xcolor}
\usepackage{array}
\usepackage{textcmds}

\title{TSBOW: Traffic Surveillance Benchmark \\for Occluded Vehicles \\Under Various Weather Conditions
    \thanks{This paper has been accepted by the 40th AAAI Conference on Artificial Intelligence (AAAI-26), \url{https://doi.org/10.1609/aaai.v40i7.37439}}
}

\date{January 22, 2026}

\author{\href{https://orcid.org/0000-0003-0662-4036}{\includegraphics[scale=0.06]{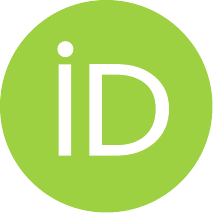}\hspace{1mm}Ngoc Doan-Minh Huynh},
	Duong Nguyen-Ngoc Tran, Long Hoang Pham, Tai Huu-Phuong Tran,\\ 
	\And 
	Hyung-Joon Jeon, Huy-Hung Nguyen, Duong Khac Vu, Hyung-Min Jeon, Son Hong Phan, \\
	\And 
	Quoc Pham-Nam Ho, Chi Dai Tran, Trinh Le Ba Khanh, \href{https://orcid.org/0000-0003-0037-112X}{\includegraphics[scale=0.06]{orcid.pdf}\hspace{1mm}Jae Wook Jeon} \\ \\
	Automation Lab, Department of Electrical and Computer Engineering \\
	Sungkyunkwan University, Suwon, South Korea \\ \\
	\texttt{\{ngochdm, jwjeon\}@skku.edu}
}

\renewcommand{\undertitle}{AAAI-26 Main Technical Track}
\renewcommand{\shorttitle}{TSBOW: Traffic Surveillance Benchmark for Occluded Vehicles Under Various Weather Conditions}

\hypersetup{
	pdftitle={TSBOW: Traffic Surveillance Benchmark for Occluded Vehicles Under Various Weather Conditions},
	pdfauthor={Ngoc D.M.~Huynh, Jae W.~Jeon},
	pdfkeywords={Computer Vision, Object Detection, Traffic Surveillance, Benchmark Dataset, Occluded Objects, Hostile Weathers},
}

\makeatletter
\def\@noticestring{%
	Copyright © 2026, Association for the Advancement of Artificial Intelligence (www.aaai.org). All rights reserved.
}
\makeatother

\begin{document}
	\maketitle
	
	\begin{abstract}
		Global warming has intensified the frequency and severity of extreme weather events, which degrade CCTV signal and video quality while disrupting traffic flow, thereby increasing traffic accident rates. Existing datasets, often limited to light haze, rain, and snow, fail to capture extreme weather conditions. To address this gap, this study introduces the \textbf{T}raffic \textbf{S}urveillance \textbf{B}enchmark for \textbf{O}ccluded vehicles under various \textbf{W}eather conditions (\textbf{TSBOW}), a comprehensive dataset designed to enhance occluded vehicle detection across diverse annual weather scenarios. Comprising over \textbf{32 hours} of real-world traffic data from densely populated urban areas, TSBOW includes more than \textbf{48,000 manually annotated} and \textbf{3.2 million semi-labeled frames}; bounding boxes spanning eight traffic participant classes from large vehicles to micromobility devices and pedestrians. We establish an object detection benchmark for TSBOW, highlighting challenges posed by occlusions and adverse weather. With its varied road types, scales, and viewpoints, TSBOW serves as a critical resource for advancing Intelligent Transportation Systems. Our findings underscore the potential of CCTV-based traffic monitoring, pave the way for new research and applications. The TSBOW dataset is publicly available at: \href{https://github.com/SKKUAutoLab/TSBOW}{https://github.com/SKKUAutoLab/TSBOW}.
	\end{abstract}

	
	\makeatletter\@notice\makeatother

	\begin{figure*}[t]
		\centering
		\includegraphics[width=\textwidth]{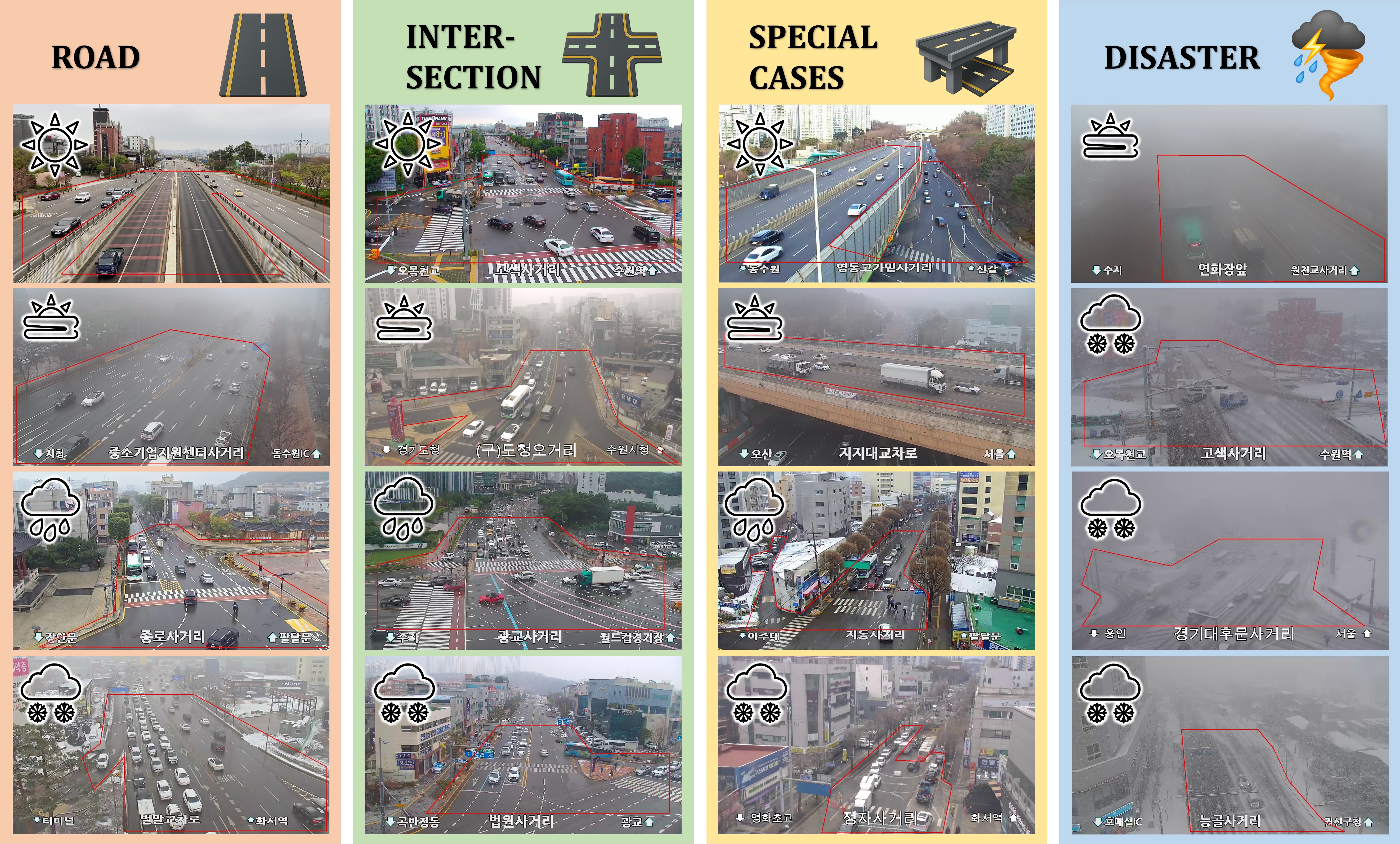}
		\caption{
			\textbf{Scenes from the TSBOW dataset}, comprising 198 videos recorded across four distinct scenarios spanning all seasons (sunny/cloudy, haze/fog, rain, snow) over a year. The dataset emphasizes adverse weather conditions and densely populated urban areas with heavy traffic, addressing significant challenges in image degradation and vehicle occlusion.
		}
		\label{fig:selection_scenes_weathers}
	\end{figure*}

	\begin{figure*}[t]
		\centering
		\includegraphics[width=\textwidth]{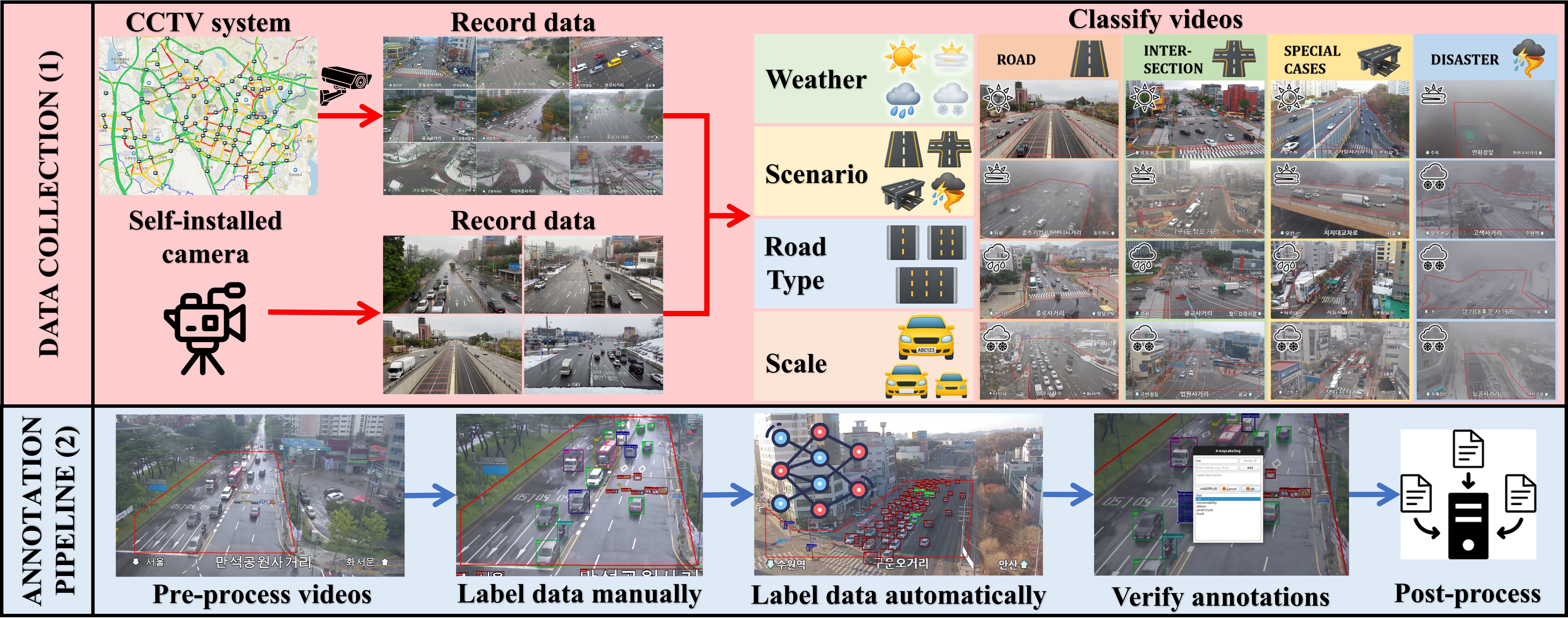}
		\caption{\textbf{Detailed overview of the data collection and annotation pipeline}. The process commences with the recording and categorization of videos during the data collection phase. Subsequently, the videos are preprocessed and allocated to a team of annotators for manual labeling. Next, a state-of-the-art model is fine-tuned to automatically annotate the remaining frames. The resulting annotations are then verified against predefined labeling criteria. Finally, the annotated instances are aggregated and undergo post-processing to finalize the dataset.}
		\label{fig:pipeline}
	\end{figure*}

	\section{Introduction}
	\label{sec:intro}
	
	Climate change has escalated the frequency and intensity of extreme weather events, significantly challenging computer vision tasks by degrading connection and image quality. These conditions disrupt traffic flow, increase traffic congestion and accident rates. 
	
	Analyzing traffic flow is essential for understanding traffic surveillance systems, enhancing transportation infrastructure through various applications. Public datasets and benchmarks have significantly advanced machine perception tasks, including image classification (\citeauthor{image_classification, image_net}), object detection (\citeauthor{object_detection, coco}), object tracking (\citeauthor{object_tracking, ade20k}), semantic segmentation (\citeauthor{segmentation, lvis}).
	
	Recent developments introduce specialized traffic datasets such as Dataset Quantization (\citeauthor{dataset_quantization}), compressed large datasets into smaller subsets; PointOdyssey (\citeauthor{pointodyssey}), designed for long-term point tracking; TrafficCAM (\citeauthor{trafficcam}), focused on traffic flow segmentation; and TUMTraf Video QA (\citeauthor{tumtraf}), targets unified spatio-temporal video understanding. 
	
	For object detection task, most existing traffic surveillance benchmarks rely on offline data captured using individual cameras, maintaining video quality in light rain or snow but proving inadequate under extreme weather conditions, such as heavy winds. Established object detection benchmarks, including UAVDT (\citeauthor{uavdt}) and UA-DETRAC (\citeauthor{uadetrac}), cover sunny and light rainy conditions but exclude severe weather scenarios.
	
	To advance traffic surveillance research, we introduce the \textbf{T}raffic \textbf{S}urveillance \textbf{B}enchmark for \textbf{O}ccluded vehicles under various \textbf{W}eather conditions (\textbf{TSBOW}), a comprehensive dataset derived from CCTV footage across diverse urban and highway routes. TSBOW encompasses a range of road types—urban streets, standard roads, and boulevards—with objects at fine, medium, and coarse scales, presenting significant challenges for detection models. Spanning a full year, the dataset captures a wide array of weather conditions, from clear skies to heavy snowfall, surpassing existing benchmarks in weather diversity (\cref{fig:selection_scenes_weathers}).
	
	Our primary contributions are outlined as follows:
	\begin{itemize}
		\item 
		Development of a semi-automatic iterative annotation pipeline for efficient and accurate labeling (\cref{fig:pipeline}).
		\item 
		Introduction of TSBOW, a novel large-scale traffic surveillance dataset comprising 198 videos and over 3.2 million extracted frames across 145 regions of interest. Collected over four seasons, TSBOW includes diverse weather conditions, notably heavy haze and snow, and covers varied road types, including straight roads, intersections, shared lanes, overpasses, and constructions.
		\item 
		Compilation of frames from densely populated areas with numerous, often occluded objects, posing significant challenges for object detection. The dataset includes diverse road types hosting eight object categories, including vehicles and pedestrians, with a balanced class distribution. Traffic lights and signs partially obscuring vehicles are also annotated, enhancing the differentiation of vehicle features from backgrounds.
		\item 
		Manual annotation and verification of a substantial portion of the data by trained personnel for training, validation, and testing. Bounding boxes were independently labeled, cross-checked for consistency. The SOTA detection model is fine-tuned to annotate remaining frames.
		\item 
		An object detection baseline for TSBOW provides a benchmark for real-time detection applications.
	\end{itemize}

	\begin{table*}[t]
		\centering
		\fontsize{7}{8}\selectfont
		\setlength{\tabcolsep}{2.1mm}
		\begin{tabular}{l|c|c|c|c|c|c|c c c|c c c c}
			
			\hline
			\multicolumn{1}{c|}{\multirow{2}{*}{\textbf{Dataset}}} & 
			\multicolumn{1}{c|}{\multirow{2}{*}{\textbf{Camera}}} & 
			\multirow{2}{*}{\textbf{\begin{tabular}[c]{@{}c@{}}Total\\ Duration\end{tabular}}} & 
			\multirow{2}{*}{\textbf{\begin{tabular}[c]{@{}c@{}}Manually\\ Labeled Fr.\end{tabular}}} & 
			\multirow{2}{*}{\textbf{\begin{tabular}[c]{@{}c@{}}Total\\ Frames\end{tabular}}} & 
			\multirow{2}{*}{\textbf{\begin{tabular}[c]{@{}c@{}}Labeling\\ Classes\end{tabular}}} & 
			\multirow{2}{*}{\textbf{Resolution}} & 
			\multicolumn{3}{c|}{\textbf{FPS}} &
			\multicolumn{4}{c}{\textbf{Unique Scenes}}
			\\
			\cline{8-10} \cline{11-14}
			& & & & & & & 20 & 25 & 30 & Sunny & Haze & Rain & Snow \\
			
			\hline
			
			UAVDT         & UAVs  & 10 hrs     & 40,409  & 80K   & 3 & 1080 $\times$ 540 & x &   &   & 42 & 1 & 7  &   \\
			UA-DETRAC     & CCTV  & 10 hrs     & --      & 140K  & 5 & 960 $\times$ 540  &   & x &   & 44 &   & 12 &   \\
			AAURainSnow   & CCTV  & 1.83 hrs   & 2,200   & 132K  & - & 640 $\times$ 480  &   &   & x &    & 1 & 5  & 16 \\
			
			\textbf{TSBOW (ours)} & CCTV  & \textbf{32.36 hrs} & \textbf{48,061} & \textbf{3.2M} & \textbf{8} & \textbf{1280 $\times$ 720} & \textbf{x} & \textbf{x} & \textbf{x} & \textbf{52} & \textbf{15} & \textbf{46} & \textbf{85} \\
			
			\hline
		\end{tabular}
		\caption{Comparison of traffic surveillance datasets.}
		\label{tab:dataset_comparison}
	\end{table*}

	\section{Related Works}
	\label{sec:relatedworks}

	\subsection{Traffic Surveillance Dataset}
	
	Traffic surveillance systems depend on high-quality, diverse datasets to optimize performance~\cite{pedestrian_tracking, object_detection_survey, enhancing_yolo}. Several public traffic datasets support this development, such as Waymo~\cite{waymo}, which provides 2D and 3D bounding boxes; TrafficMOT~\cite{trafficmot}, focused on multi-object tracking; eTram~\cite{etram}, offering 2D bounding boxes for event-based cameras; STEP~\cite{STEP}, providing object segmentation and tracking; and OVT-B~\cite{OVTB}, a benchmark for vocabulary multi-object tracking.

	\begin{figure}[t]
		\centering
		\includegraphics[width=0.5\columnwidth]{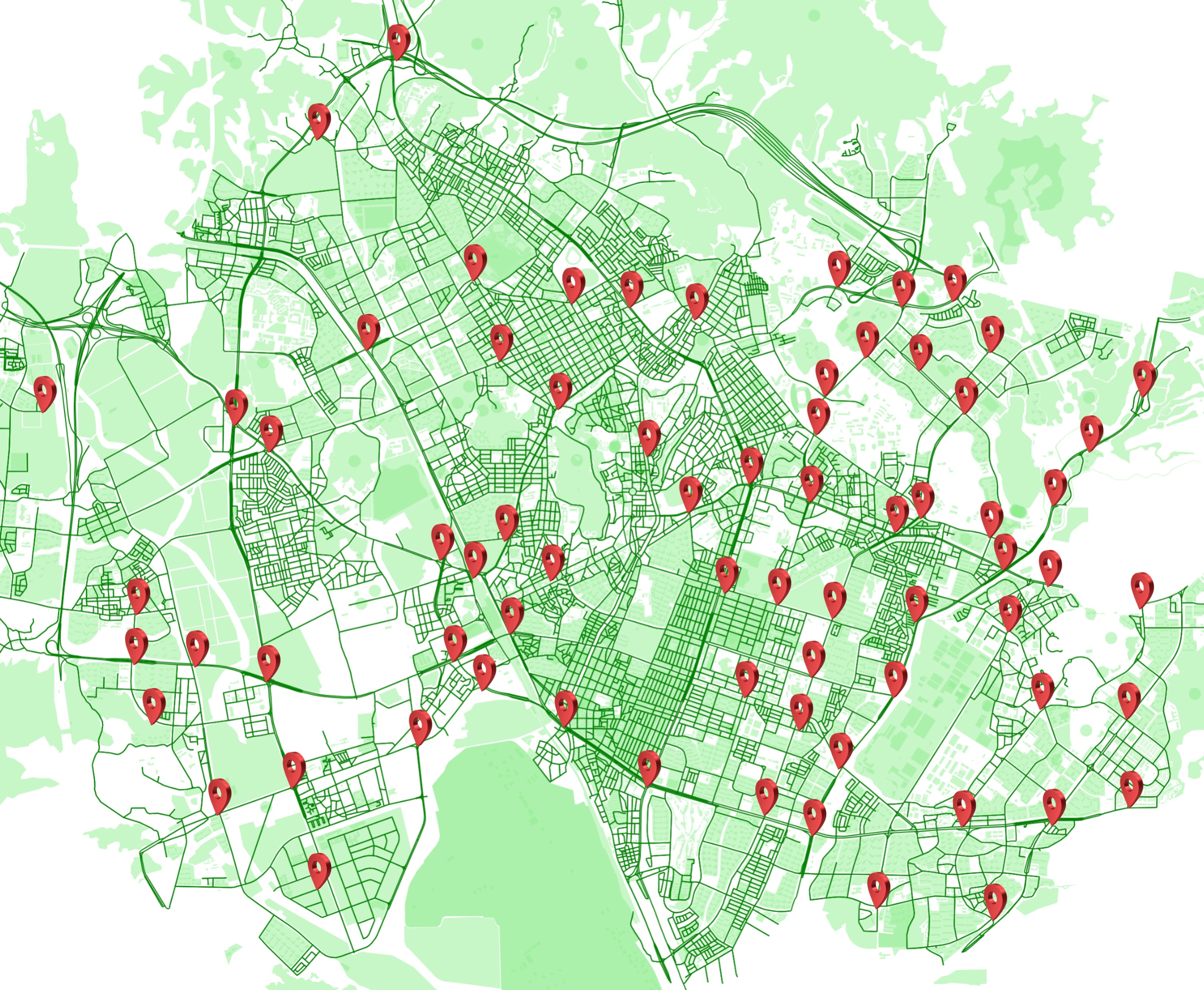}
		\caption{Suwon recording locations in TSBOW dataset.}
		\label{fig:collection_routes}
	\end{figure}
	
	\begin{figure}[t]
		\centering
		\includegraphics[width=0.75\columnwidth]{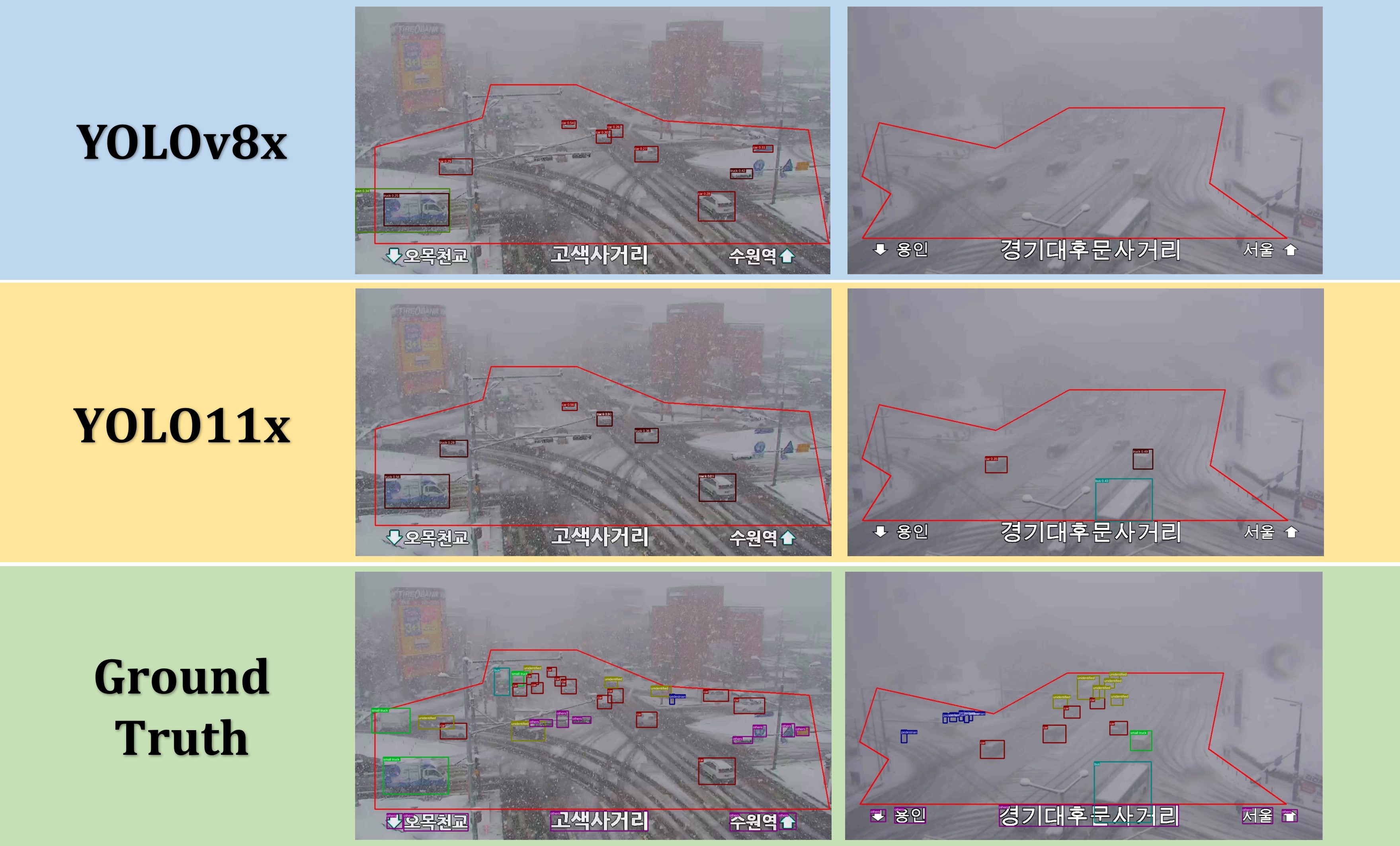}
		\caption{An example of detecting vehicles in heavy snow by using
			the YOLOv8x and YOLO11x models}
		\label{fig:detection_examples}
	\end{figure}

	Recent advancements in autonomous driving frequently combine data from color cameras and LiDAR sensors. Color cameras capture visual details, such as color, texture, and semantic information, facilitating the creation of 2D bounding boxes. In contrast, LiDAR generates 3D spatial data, including distance, depth, and point clouds. Notable autonomous driving datasets include Ithaca365~\cite{Ithaca365} and SODA10M~\cite{SODA10M}, which provide 3D bounding boxes; HoloVIC~\cite{HoloVIC}, offering 3D bounding boxes and multi-object tracking; and TAP-Vid~\cite{TAPVid}, which includes object tracking points in videos.
	Despite their complementary strengths, the integration of camera and LiDAR data presents notable limitations. First, LiDAR's performance is limited by its height and coverage range, particularly when positioned at elevated locations, resulting in unreliable data. Second, LiDAR struggles to detect small objects, such as pedestrians and bicycles, at long distances due to sparse point clouds, which provide limited actionable information. Third, many existing traffic surveillance systems rely exclusively on color cameras, as incorporating LiDAR is often cost-prohibitive and impractical. Consequently, this research leverages existing government CCTV systems to analyze traffic flow across diverse weather conditions over a year, with a particular focus on disasters that significantly disrupt traffic.
	
	The UAVDT dataset~\cite{uavdt} comprises 10 hours of UAV-captured video across urban areas under sunny and rainy conditions. 
	This dataset presents detection challenges, including water puddle reflections, shadows, and camera motion blur, exacerbated by UAV altitudes ranging from low to high (above 70 meters), rendering it less applicable to ground-based surveillance like CCTV systems. Conversely, UA-DETRAC~\cite{uadetrac} includes 10 hours of video recorded with a Canon camera~\cite{canon} across 24 locations in China under four weather conditions (cloudy, nighttime, sunny, rainy). While sharing similar challenges—water puddles, shadows, and motion blur—UA-DETRAC’s ground-proximate camera setup enhances model training performance but reduces complexity and real-world representativeness compared to UAVDT.
	
	Both UAVDT and UA-DETRAC datasets are limited to sunny and rainy conditions, overlooking snowfall, a critical factor affecting video quality. Addressing this gap, the AAU RainSnow Traffic Surveillance Dataset~\cite{aaurainsnow} is introduced, captured using both a conventional RGB color camera and a thermal infrared camera. Comprising 22 five-minute videos, this dataset documents rainfall and snowfall across seven intersections in Denmark, providing segmentation for 13,297 objects under four weather conditions: rain, snow, haze, and fog, though it omits bounding boxes for vehicles. The AAU RainSnow dataset exhibits shared challenges with UAVDT and UA-DETRAC, including puddle reflections, raindrops on the lens, and camera variations. Like UA-DETRAC, its ground-proximate camera positioning reduces complexity, limiting its applicability to real-world traffic surveillance systems such as CCTV-based monitoring.
	
	As shown in Tab.~\ref{tab:dataset_comparison}, compared to others, our dataset features longer video recordings and higher-quality traffic videos with higher-resolution frames. Additionally, it offers greater diversity in FPS, weather conditions, and scenarios compared to other benchmark datasets. Specifically, we include special scenarios and disaster cases that have not been covered in previous datasets. Because of limiting resources, we first focus on different weathers in day time, the night time will be updated in subsequent versions of our dataset.
	
	\noindent
	\subsection{Object detection}
	
	Object detection~\cite{object_detection_survey} is a machine learning task that involves image localization and object classification. 
	Several high-accuracy object detection models have been developed, such as Faster R-CNN~\cite{fastrcnn}, CenterNet~\cite{centernet}, DETR (DEtection TRansformer)~\cite{detr}. However, when balancing processing speed and accuracy, the YOLO family of models emerges as a promising choice due to its exceptional performance and real-time processing capabilities. Various versions of YOLO have been proposed~\cite{YOLOversions, YOLOversions2} are introduced, such as YOLOv3~\cite{yolov3}, YOLOv5~\cite{yolov5}, YOLOv8~\cite{yolov8}, YOLOX~\cite{yoloX}, YOLOv11~\cite{yolov11}, and YOLOv12~\cite{yolov12}. Compared to earlier versions, YOLOv8 features an enhanced backbone and employs a path aggregation network, achieving high accuracy while maintaining real-time processing. The YOLOv11 model further integrates vision transformers (ViTs)~\cite{vits} to enhance contextual understanding, yielding the highest accuracy in the mean Average Precision (mAP) metric, albeit with a slight reduction in processing speed. YOLOv12 combines FlashAttention and an R-ELAN backbone to achieve higher accuracy without compromising real-time detection. RT-DETR~\cite{rtdetr}, leveraging a transformer architecture, excels in dense and complex scene understanding. 
	In this study, we employ YOLOv8, YOLOv11, YOLOv12, and RT-DETR, specifically the x-large variants, to establish experimental baselines for the \textbf{T}raffic \textbf{S}urveillance \textbf{B}enchmark for \textbf{O}ccluded vehicles under various \textbf{W}eather conditions (\textbf{TSBOW}).

	
	\section{CCTV Traffic Surveillance Benchmark}
	\label{sec:dataset}
	
	The Traffic Surveillance Benchmark for Occluded vehicles under various Weather conditions (TSBOW) dataset is specifically engineered to capture traffic flow within the diverse road networks of Suwon city, Gyeonggi, Korea.

	\begin{figure}[t]
		\centering
		\includegraphics[width=0.7\columnwidth]{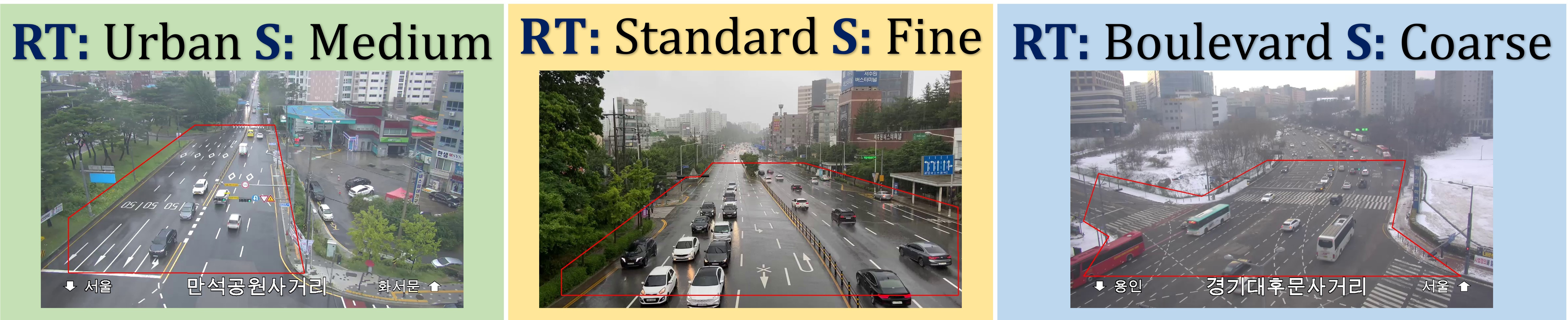}
		\caption{An example of road types (RT) and scales (S)}
		\label{fig:example_roadtype_scales}
	\end{figure}

	\begin{figure}[t]
		\centering
		\includegraphics[width=0.7\columnwidth]{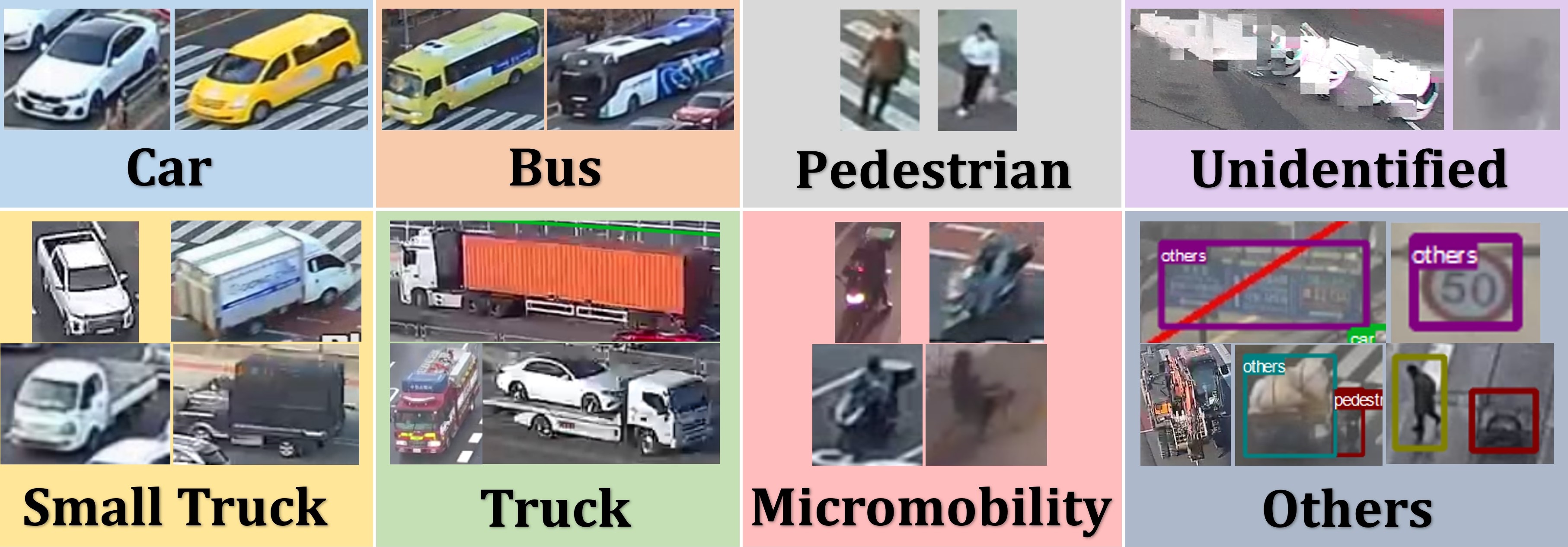}
		\caption{Visualization of annotated instances of different classes in TSBOW dataset.}
		\label{fig:vehicle_categories}
	\end{figure}

	\begin{figure}[t]
		\centering
		\includegraphics[width=0.75\columnwidth]{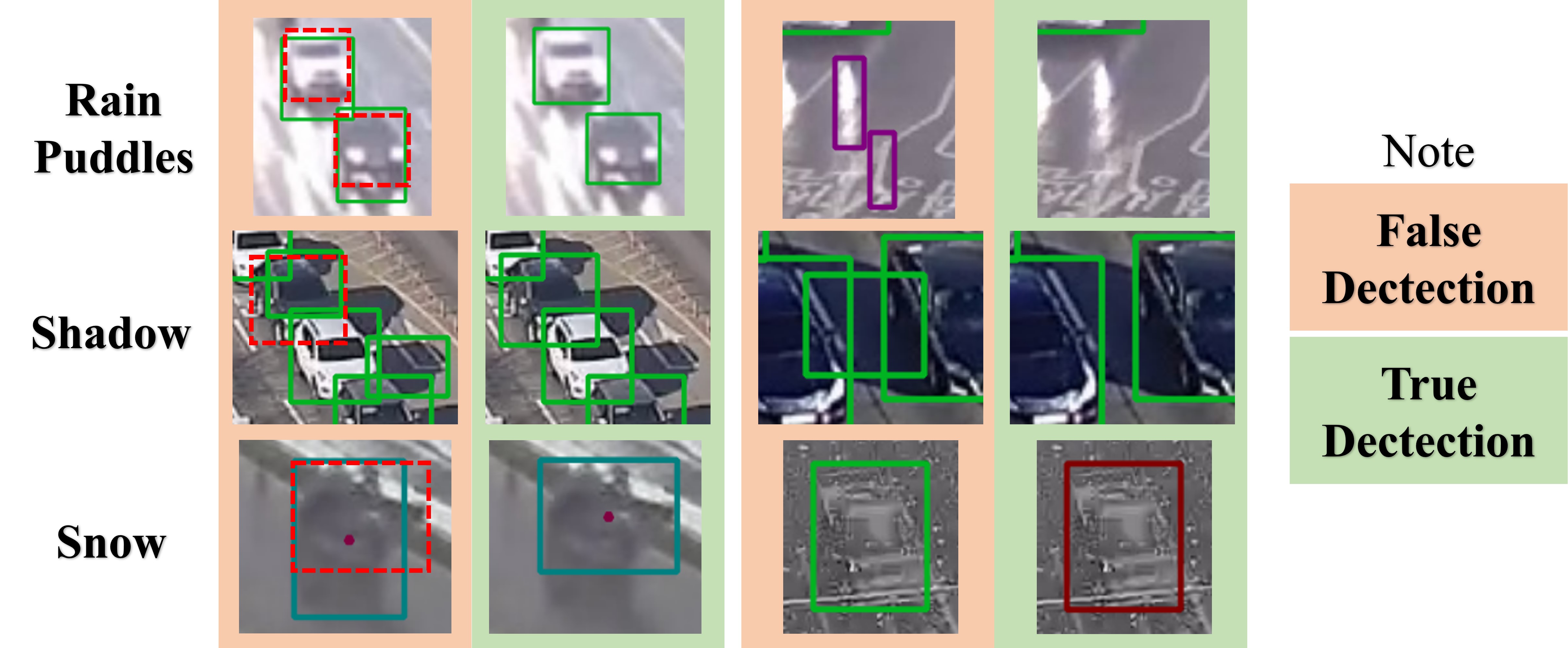}
		\caption{Challenges by Weather Conditions in TSBOW.}
		\label{fig:data_challenges}
	\end{figure}
	
	\begin{table*}[t]
		\centering
		\fontsize{8}{9}\selectfont
		\setlength{\tabcolsep}{9mm}
		\begin{tabular}{l|c|c|c|c|c}
			\hline
			\textbf{Scenario/Weather} & \textbf{Normal} & \textbf{Haze} & \textbf{Rain} & \textbf{Snow} & \textbf{Total} \\
			\hline
			Road            & 12     & 6    & 15   & 28   & \textbf{61} \\
			Intersection    & 28     & 6    & 25   & 37   & \textbf{96} \\
			Special cases   & 12     & 2    & 6    & 11   & \textbf{31} \\
			Disaster        & --     & 1    & --   & 9    & \textbf{10} \\
			\hline
			\textbf{Total}           & \textbf{52}     & \textbf{15}   & \textbf{46}   & \textbf{85}   & \textbf{198} \\
			\hline
		\end{tabular}
		\caption{Statistics on scenarios and weathers in TSBOW.}
		\label{tab:tsbow_stats_scenario_weather}
	\end{table*}

	\noindent
	\subsection{Collection Routes} 
	
	Our dataset is derived from fixed routes comprising a wide range of scenes (\cref{fig:collection_routes}), enabling robustness evaluation under diverse weather conditions. It is systematically classified according to distinct attributes, including scenario, weather, road type, and scale. Detailed descriptions are provided below.
	
	First, the video scenarios are categorized into four distinct types—road, intersection, special case, and disaster—as illustrated in \cref{fig:selection_scenes_weathers}. 
	\begin{itemize}
		\item 
		\textbf{Road} comprises straight roads or those where traffic flow remains unaffected by traffic lights.
		\item 
		\textbf{Intersection} includes three subtypes—T, Y, or crossroad—and extends to scenes where traffic lights or pedestrian crossings influence traffic flow, thereby warranting classification as intersections.
		\item 
		\textbf{Special case} includes videos featuring shared lanes, overpasses, or mid-road construction. Shared lanes, including narrow one-way variants, are characterized by bidirectional traffic and pedestrian activity within a single lane, prevalent in space-constrained, densely populated areas. Overpass footage is subdivided into two groups: scenes solely depicting the overpass and those capturing both the overpass and adjacent or underlying roads. The latter poses greater detection challenges due to significant scale disparities among vehicles within a single frame.
		\item 
		\textbf{Disaster} pertains to scenarios where hostile weather severely degrades video quality, such as heavy snow, rendering vehicle identification exceedingly difficult and presenting the most formidable challenge for detection models. Fig.~\ref{fig:detection_examples} illustrates the object detection outcomes using YOLOv8x and YOLOv11x. 
	\end{itemize}

	Second, videos for the TSBOW dataset, covering roads, intersections, and special cases, were recorded in Suwon under diverse weather conditions—normal (sunny), haze, rain, and snow—throughout the year. Unlike prior datasets where \qq{rain} videos resemble sunny conditions due to wet roads without visible raindrops, TSBOW classifies \qq{rain} only when raindrops or active rainfall are evident. Similarly, \qq{snow} videos require visible snowflakes or snowfall, altering object appearances (e.g., vehicles with white pixels from snow or frost). These conditions, combined with unstable connections and camera vibrations from strong winds, degrade video quality and complicate computer vision tasks. Videos with wet roads or residual snow lacking active precipitation are classified as \qq{normal} (sunny).
	In the disaster scenario, heavy haze and extreme snow significantly impair object detection by obscuring visual features, as shown in \cref{fig:detection_examples}. Snow-covered vehicles blend into white snowy backgrounds, challenging model performance and object-background differentiation. Termed \qq{disaster} due to its severe impact on traffic flow, this scenario exacerbates congestion and accident rates.

	Third, the TSBOW dataset classifies data collection zones by the number of straight lanes per direction, excluding turning lanes, into three road types: urban (two lanes, primarily cars, small trucks, and pedestrians), standard (four lanes, including larger vehicles like trailer trucks), and boulevard (over six lanes, featuring flatbed trucks and car transporters). Unlike other datasets limited to urban and standard roads, TSBOW includes boulevards, where high vehicle density and occlusion intensify detection challenges due to frequent object overlap.

	Final, CCTV cameras along routes and intersections vary in angle and height, producing bounding box sizes categorized into three scales: fine (near road surface, detailed object visualization), medium (elevated cameras, discernible license plates but reduced clarity), and coarse (distant cameras, high vehicle count but partial visibility due to distance and occlusion). Fig.~\ref{fig:example_roadtype_scales} provides an example of road types and scales. While UA-DETRAC covers fine and medium scales, UAVDT focuses on medium and coarse scales.
	
	The diverse attributes of our dataset—spanning locations (\cref{fig:collection_routes}), weather conditions, road types, and object scales (\cref{fig:example_roadtype_scales})—render it a robust resource for assessing and enhancing traffic surveillance systems. Furthermore, it incorporates eight labeling classes, surpassing other benchmark datasets in granularity, facilitating detailed classification of vehicles and pedestrians and fostering deeper insights into traffic dynamics for infrastructure improvement. Beyond object detection, the TSBOW dataset supports additional applications, including crowd counting~\cite{crowd_counting}, speed estimation~\cite{speed_estimation}, and object tracking~\cite{object_tracking_2}, thereby offering practical utility for real-world traffic management systems.

	\noindent
	\subsection{Labeling Process}
	
	The annotation process for the TSBOW dataset ensures high-quality ground truth data through five phases: video pre-processing, manual labeling, automatic labeling, annotation verification, and post-processing.
	Firstly, in pre-processing, regions of interest (ROIs) are defined to capture the main road sections where traffic objects are most visible. After that, frames are then extracted at set intervals and manually annotated using X-Anylabeling \cite{X-AnyLabeling}, an open-source tool for precise bounding box creation, focusing on vehicles and pedestrians in high-density urban roads and intersections.
	Subsequently, a YOLOv12x model trained on region-specific vehicle characteristics (e.g., size, shape, color) in Korea, is used for semi-automatic labeling of remaining frames. Annotations undergo rigorous review and quality control to eliminate substandard entries, ensuring high-quality data.
	Lastly, the annotated images are compiled and subjected to post-processing to produce the final version of the dataset.
	
	Objects within the TSBOW dataset are classified into eight distinct categories: \textit{car}, \textit{bus}, \textit{truck}, \textit{small truck}, \textit{micromobility}, \textit{pedestrian}, \textit{unidentified}, and \textit{others}. Annotated examples across these categories are depicted in \cref{fig:vehicle_categories}.
	
	In the first version of the dataset, which is aimed for public release, 48,061 frames have been manually labeled and verified. The annotations for remaining frame from over 3.2M were generated using the YOLOv12x model. Subsequent versions will substantially increase the proportion of manually annotated frames. License plates and pedestrian faces have been obscured to comply with privacy regulations.

	\begin{table*}[t]
		\centering
		\fontsize{8}{9}\selectfont
		\setlength{\tabcolsep}{11mm}
		\begin{tabular}{l|c|c|c|c}
			\hline
			\textbf{Road Type/Scale} & \textbf{Fine} & \textbf{Medium} & \textbf{Coarse} & \textbf{Total} \\
			\hline
			Urban      & 14 & 66 & 9  & \textbf{89} \\
			Standard   & 13 & 61 & 4  & \textbf{78} \\
			Boulevard  & 2  & 20 & 9  & \textbf{31} \\
			\hline
			\textbf{Total}      & \textbf{29} & \textbf{147} & \textbf{22} & \textbf{198} \\
			\hline
		\end{tabular}
		\caption{Statistics on scales and road types in the TSBOW.}
		\label{tab:tsbow_stats_road_scale}
	\end{table*}

	\begin{table*}[t]
		\centering
		\fontsize{8}{9}\selectfont
		\setlength{\tabcolsep}{8mm}
		\begin{tabular}{l|c|c|>{\columncolor{blue!10}}c|>{\columncolor{blue!10}}c}
			\hline
			\textbf{Instances} & 
			\textbf{\begin{tabular}[c]{@{}c@{}}UAVDT\\ Ground Truth\end{tabular}} & 
			\textbf{\begin{tabular}[c]{@{}c@{}}UA-DETRAC\\ Ground Truth\end{tabular}} & 
			\textbf{\begin{tabular}[c]{@{}c@{}}TSBOW\\ Ground Truth\end{tabular}} & 
			\textbf{\begin{tabular}[c]{@{}c@{}}Manually\\ Labeled\end{tabular}} \\
			\hline
			Car             & 756{,}166  & 1{,}052{,}408  & 50{,}269{,}953   & 783{,}203 \\
			Bus             & 0          & 95{,}570       & 2{,}252{,}616    & 35{,}715 \\
			Truck           & 0          & 20{,}641       & 565{,}932        & 10{,}657 \\
			Small Truck     & -          & 105{,}436      & 3{,}258{,}031    & 58{,}272 \\
			Pedestrian      & -          & -              & 2{,}935{,}102    & 53{,}414 \\
			Micromobility   & -          & -              & 1{,}691{,}744    & 29{,}959 \\
			Unidentified    & -          & -              & 515{,}447        & 8{,}379 \\
			Others          & 0          & -              & 9{,}648{,}223    & 151{,}556 \\
			\hline
			\textbf{Total}  & \textbf{756{,}166} & \textbf{1{,}274{,}055} & \textbf{71{,}137{,}048} & \textbf{1{,}131{,}155} \\
			\hline
		\end{tabular}
		\caption{Instance statistics across UAVDT, UA-DETRAC, and TSBOW datasets.}
		\label{tab:annotated_instances}
	\end{table*}

	\noindent
	\subsection{Dataset Statistic and Characteristics}
	
	This study analyzes data statistics under two conditions: contextual influences and road structures. Tab.~\ref{tab:tsbow_stats_scenario_weather} details video counts by scenario and weather, with intersections—impacted by traffic lights and pedestrians—exhibiting more occluded objects, prompting increased data collection. Video distribution across scenes remains balanced, comparable to other scenarios.
	Tab.~\ref{tab:tsbow_stats_road_scale} categorizes videos by scale and road type, with urban and standard roads prevailing over boulevards due to the city-zone focus. CCTV cameras, primarily offering medium-scale perspectives, are strategically placed to capture traffic dynamics, with sufficient fine- and coarse-scale videos to cover diverse scenarios.
	
	As previously noted, our dataset comprises over 3.2 million frames across 198 videos. Tab.~\ref{tab:annotated_instances} details the bounding box counts for each class in three datasets: UAVDT, UA-DETRAC, and our TSBOW. Of the 1.1 million manually annotated objects, cars constitute 69\%, with camera information and traffic signs contributing to the proportion of other objects. Unlike benchmark datasets, where cars exceed 83\%, our dataset achieves a more balanced distribution across eight classes. The high prevalence of pedestrians and micromobility devices indicates recordings from densely populated urban areas. Frames average 24 objects, with a maximum of 122. The highest number of semi-labeled bounding boxes in a single video reaches 1,233,828 across 17,789 frames, reflecting frequent object occurrences. 
	Bounding boxes are classified by occlusion level, measured as the percentage of area occluded: no occlusion ($<$15\% IoU), light occlusion (15--$<$40\% IoU), and heavy occlusion ($\geq$40\% IoU). Their distribution is 721,684 (no occlusion), 266,420 (light occlusion), and 143,051 (heavy occlusion). Accordingly, traffic flow is categorized into light (44 videos), moderate (98 videos), and heavy (56 videos).
	This substantial instance count and balanced class distribution enhance the dataset’s reliability for traffic surveillance research.

	In our TSBOW dataset, numerous factors challenge detection models in accurately identifying and localizing objects, as outlined below:
	\begin{itemize}
		\item 
		Weather conditions: Normal weather scenarios—cloudy and sunny—impact detection performance. Cloudy conditions eliminate vehicle shadows on road surfaces, simplifying detection. In contrast, sunny conditions introduce shadows, leading to less precise bounding boxes that encompass both vehicles and their shadows, reducing model accuracy (\cref{fig:data_challenges}). Haze degrades image quality, obscuring object features and further complicating detection. Rainy conditions create water puddles that distort bounding box sizes, while strong winds and unstable connections during rain or snow induce camera movement, causing motion blur. 
		\item 
		Scenarios: Continuous traffic flow on roads and overpasses enhances detection model performance due to minimal vehicle occlusions. However, simultaneous recording of overpasses and underlying or adjacent roads complicates detection due to varying vehicle scales and viewing angles. Conversely, traffic lights disrupt flow, causing significant occlusions where only partial objects are visible, posing substantial challenges for models in localizing vehicles and often failing to detect highly occluded instances. Road construction further exacerbates occlusions by reducing lanes, leading to traffic congestion. These issues, compounded by disaster scenarios, underscore the detection difficulties discussed.
	\end{itemize}
	These characteristics collectively impede the precision and reliability of object detection models across diverse environmental and situational contexts. \textbf{More detailed descriptions are mentioned in the Supplementary material}. 
	

	\section{Experiments}
	\label{sec:experiments}
	
	\subsection{Object Detection Qualitative Result}
	
	In this section, we establish benchmarks for the TSBOW dataset using neural network-based object detection methods. We utilize YOLOv8x, YOLO11x, YOLOv12x, and RT-DETR-x models, pretrained on the COCO dataset~\cite{coco} and fine-tuned on our dataset at an image resolution of 1280 pixels, as lower resolutions impair inference performance, particularly for occluded vehicles. Evaluation metrics include average precision (AP), mean average precision (mAP), intersection over union (IoU), precision, and recall.
	
	Each video is segmented into three subsets: the first 5 minutes for testing, the next 2 minutes for validation, and the final 3 minutes for training, with frame details provided in Tab.~\ref{tab:frame_statistics}. To ensure reliable model performance, manually labeled sets are used for training, validation, and testing. Inference parameters include an IoU threshold of 0.6, an image size of 1280 pixels, and a confidence score of 0.5. 
	
	\begin{table*}[t]
		\centering
		\fontsize{8}{9}\selectfont
		\setlength{\tabcolsep}{9mm}
		\begin{tabular}{l|c|c|c|c}
			\hline
			\textbf{Subset} & \textbf{Total} & \textbf{Training} & \textbf{Validation} & \textbf{Test} \\
			\hline
			Manually labeled frames & 48{,}061 & 10{,}881 & 7{,}559 & 29{,}621 \\
			Total frames            & 3{,}267{,}598 & 973{,}877 & 658{,}856 & 1{,}634{,}865 \\
			\hline
		\end{tabular}
		\caption{Statistics of manually labeled and total video frames across training, validation, and test subsets.}
		\label{tab:frame_statistics}
	\end{table*}

	\begin{table*}[t]
		\centering
		\fontsize{8}{9}\selectfont
		\setlength{\tabcolsep}{11mm}
		\begin{tabular}{c|c|c|c|c}
			\hline
			\textbf{Method} & \textbf{Precision} & \textbf{Recall} & \textbf{mAP50} & \textbf{mAP50--95} \\
			\hline
			YOLOv8x    & 0.783 & 0.705 & 0.733 & 0.609 \\
			YOLO11x    & 0.786 & 0.696 & 0.734 & 0.614 \\
			YOLOv12x   & \textbf{0.806} & 0.662 & \textbf{0.744} & \textbf{0.615} \\
			RT-DETR-x  & 0.731 & \textbf{0.740} & 0.718 & 0.552 \\
			\hline
		\end{tabular}
		\caption{Model performances after training 100 epochs and validating with imgsz=1280 \textbf{on manually labeled test set}.} 
		\label{tab:model_performance_imgsz1280}
	\end{table*}

	\begin{figure}[t]
		\centering
		\includegraphics[width=0.75\columnwidth]{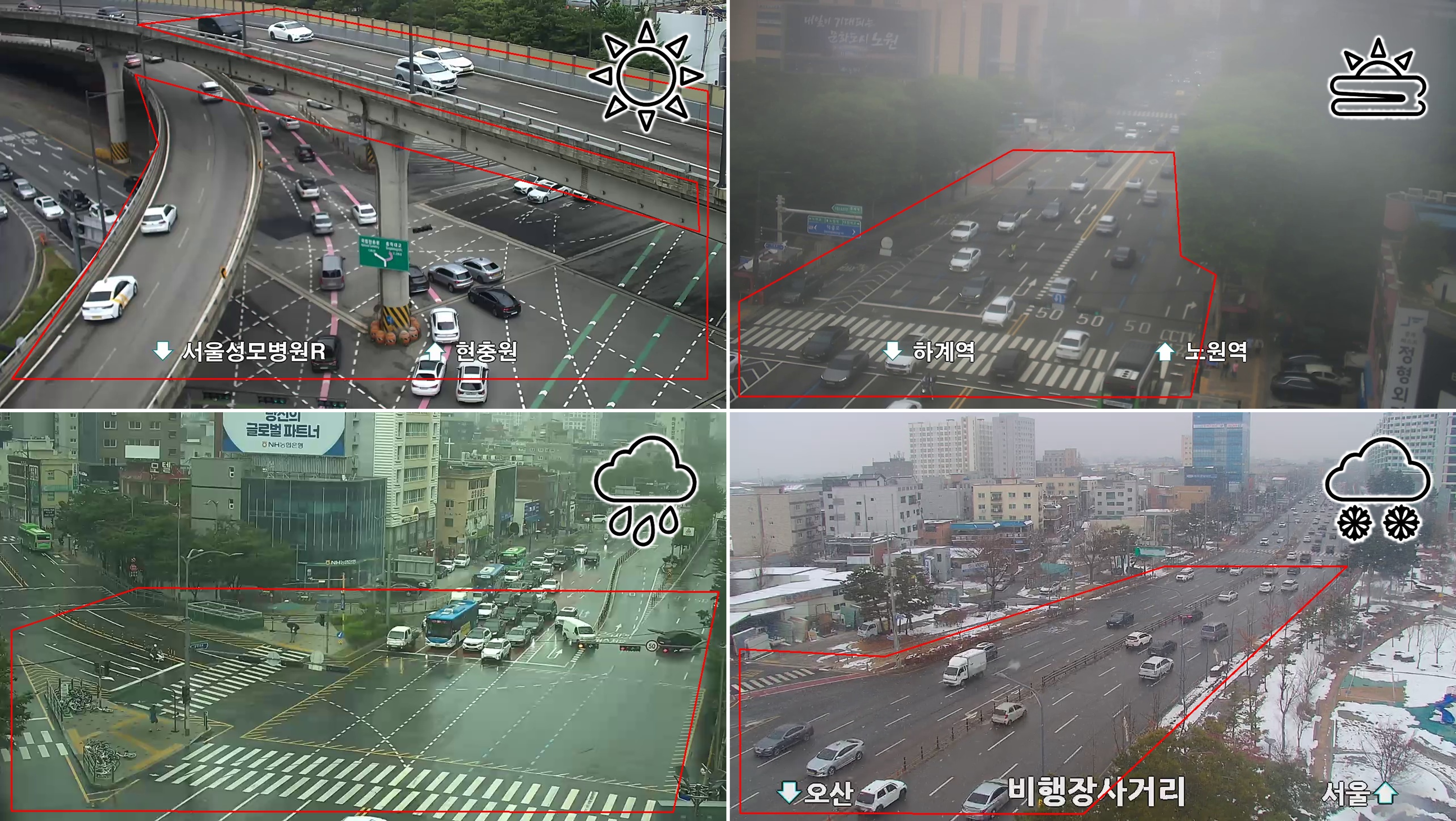}
		\caption{Selected scenes for comparison with other datasets}
		\label{fig:Comparison}
	\end{figure}

	\begin{table*}[t]
		\centering
		\fontsize{8}{9}\selectfont
		\setlength{\tabcolsep}{8mm}
		\begin{tabular}{l|c|c|c|c}
			\hline
			\textbf{Method} & \textbf{Precision} & \textbf{Recall} & \textbf{mAP50} & \textbf{mAP50-95} \\
			\hline
			YOLOv12x trained on \textbf{UAVDT} & {0.647} & {0.141} & {0.383} & {0.328} \\
			YOLOv12x trained on \textbf{UA-DETRAC} & {\textbf{0.820}} & {0.295} & {0.558} & {0.459} \\
			YOLOv12x trained on \textbf{TSBOW} (ours) & {0.743} & {\textbf{0.869}} & {\textbf{0.846}} & {\textbf{0.792}} \\
			\hline
		\end{tabular}
		\caption{Models performance for \textit{car} across different metrics \textbf{on the comparison set}.}
		\label{tab:comparison_set}
	\end{table*}
	
	\begin{table*}[t]
		\centering
		\fontsize{8}{9}\selectfont
		\setlength{\tabcolsep}{6.1mm}
		\begin{tabular}{l|c|c|c|c|c|c}
			\hline
			\textbf{Class} & \textbf{Images} & \textbf{Instances} & \textbf{Precision} & \textbf{Recall} & \textbf{mAP50} & \textbf{mAP50--95} \\
			\hline
			\textbf{All}  & \textbf{29,621} & \textbf{689,839} & \textbf{0.806} & \textbf{0.662} & \textbf{0.744} & \textbf{0.615} \\
			Unidentified  & 2,177   & 4,855   & 0.475 & 0.223 & 0.317 & 0.221 \\
			Others        & 14,248  & 90,276  & 0.769 & 0.628 & 0.696 & 0.619 \\
			Pedestrian    & 11,807  & 32,779  & 0.833 & 0.605 & 0.715 & 0.447 \\
			Micromobility & 9,124   & 18,490  & 0.793 & 0.574 & 0.726 & 0.519 \\
			Car           & 29,244  & 479,560 & 0.959 & 0.932 & 0.959 & 0.849 \\
			Bus           & 12,378  & 21,037  & 0.917 & 0.929 & 0.951 & 0.876 \\
			Small truck   & 16,176  & 36,152  & 0.878 & 0.830 & 0.870 & 0.750 \\
			Truck         & 5,390   & 6,690   & 0.824 & 0.575 & 0.720 & 0.643 \\
			\hline
		\end{tabular}
		\caption{YOLOv12x performance across different classes.}
		\label{tab:yolov12x_class_performance}
	\end{table*}
	
	\begin{table*}[t]
		\centering
		\fontsize{8}{9}\selectfont
		\setlength{\tabcolsep}{7.9mm}
		\begin{tabular}{ll|c|c|c|c}
			\hline
			\textbf{Group} & \textbf{Category} & \textbf{Precision} & \textbf{Recall} & \textbf{mAP50} & \textbf{mAP50--95} \\
			\hline
			\multirow{4}{*}{\textbf{Scenario}} 
			& Road         & 0.756 & 0.608 & 0.687 & 0.575 \\
			& Intersection & \textbf{0.827} & 0.677 & \textbf{0.759} & \textbf{0.633} \\
			& Special cases & 0.800 & \textbf{0.681} & 0.747 & 0.615 \\
			& Disaster     & 0.765 & 0.559 & 0.656 & 0.510 \\
			\hline
			\multirow{4}{*}{\textbf{Weather}} 
			& Normal & 0.791 & 0.622 & 0.726 & 0.605 \\
			& Haze   & 0.799 & \textbf{0.732} & 0.785 & \textbf{0.684} \\
			& Rain   & \textbf{0.851} & 0.721 & \textbf{0.789} & 0.664 \\
			& Snow   & 0.783 & 0.641 & 0.723 & 0.597 \\
			\hline
			\multirow{3}{*}{\textbf{Scale}} 
			& Fine   & 0.733 & 0.619 & 0.686 & 0.559 \\
			& Medium & \textbf{0.810} & 0.666 & \textbf{0.751} & \textbf{0.630} \\
			& Coarse & 0.791 & \textbf{0.676} & 0.733 & 0.581 \\
			\hline
			\multirow{3}{*}{\textbf{Road Type}} 
			& Urban     & 0.821 & \textbf{0.675} & \textbf{0.757} & \textbf{0.632} \\
			& Standard  & \textbf{0.828} & 0.661 & 0.754 & 0.622 \\
			& Boulevard & 0.745 & 0.664 & 0.702 & 0.579 \\
			\hline
			\multirow{3}{*}{\textbf{Traffic}} 
			& Light     & 0.792 & 0.658 & 0.726 & 0.617 \\
			& Moderate  & 0.795 & 0.647 & 0.734 & 0.606 \\
			& Heavy     & \textbf{0.841} & \textbf{0.678} & \textbf{0.769} & \textbf{0.640} \\
			\hline
			
		\end{tabular}
		\caption{Influence of dataset characteristics on object detection performance}
		\label{tab:yolov12x_characteristics}
	\end{table*}

	Tab.~\ref{tab:model_performance_imgsz1280} illustrates the precision, recall, mAP50, and mAP50-95 scores of the YOLOv8x, YOLO11x, YOLOv12x, and RT-DETR-x models after training for 100 epochs. 
	In the evaluation, RT-DETR-x achieves the highest recall, prioritizing broad object coverage, but exhibits lower precision and mAP scores, indicating weaker localization performance. Conversely, YOLOv12x outperforms others in precision, mAP50, and mAP50-95, attributed to its reduced false positive rate. Thus, YOLOv12x demonstrates superior robustness for general object detection, and was selected to annotate the remaining frames.

	\noindent
	\subsection{Datasets Comparison}
	
	To ensure a fair comparison, we created a subset of medium-scale scenes distinct from the TSBOW dataset, featuring unique road structures and vehicle characteristics. While snowy conditions were recorded in Suwon, additional videos capturing normal, haze, and rain conditions were collected in Seoul (Fig.~\ref{fig:Comparison}). Unlike UA-DETRAC, which includes only fine- and medium-scale videos captured by a color camera, and UAVDT, which focuses on medium- and coarse-scale drone footage, TSBOW encompasses fine, medium, and coarse scales. Therefore, the comparison subset comprises medium-scale scenes, included in the UAVDT, UA-DETRAC, and TSBOW datasets.
	
	Tab.~\ref{tab:comparison_set} details detection performance for the \textit{car} class across various metrics on this comparison set. YOLOv12x models were trained on UAVDT, UA-DETRAC, and TSBOW datasets with identical setups. The UAVDT-trained model yielded the lowest scores, as its high-altitude drone footage is less applicable to ground-based CCTV surveillance. The UA-DETRAC-trained model achieved high precision but low recall, mAP50, and mAP50-95, due to its emphasis on clear vehicle features at specific distances, overlooking distant vehicles. Conversely, TSBOW mitigates these limitations by incorporating vehicles across diverse scales and optimizing region-of-interest (ROI) settings to enhance detection. Consequently, the TSBOW-trained model balances precision and recall, achieving superior performance in recall, mAP50, and mAP50-95.
	
	\noindent
	\subsection{Ablation Study on Object Classes}
	Tab.~\ref{tab:yolov12x_class_performance} presents the detailed detection performance of the fine-tuned YOLOv12x model across various object classes. The class \qq{car,} with high occurrence, achieves the highest scores in three of four metrics, while \qq{bus} excels in mAP50-95 due to the distinct features of fixed-shape objects. For smaller objects, such as \qq{pedestrians} and \qq{micromobility}, the model demonstrates promising detection performance.
	
	\noindent
	\subsection{Ablation Study on Data Characteristics}
	The fine-tuned YOLOv12x model is evaluated across diverse data characteristics, including weather, scenario, scale, road type, and traffic. Tab.~\ref{tab:yolov12x_characteristics} provides detailed performance metrics, including precision, recall, mAP50, and mAP50-95. In the \qq{disaster} scenario, heavy snow significantly obscures object features, markedly impairing detection. Under weather conditions, \qq{normal} yield lower scores than \qq{rain} due to frequent vehicle overlap. Similarly, the \qq{coarse} scale, characterized by numerous small and heavily occluded objects, poses significant detection challenges. For road type, \qq{boulevards,} with high vehicle density and occlusion, present substantial obstacles to detection accuracy. Object detectors often struggle with heavily occluded objects, frequently misidentifying two to three occluded vehicles as a single object, leading to numerous missed detections.

	\section{Conclusion and Future Works}
	\label{sec:conclusion}
	
	This study introduces the Traffic Surveillance Benchmark for Occluded vehicles under various Weather conditions (TSBOW), a comprehensive, semi-automatically annotated traffic surveillance dataset designed to improve monitoring system training, particularly under extreme weather conditions such as heavy haze and snow. Collected across all seasons and diverse road scenarios, TSBOW comprises 32 hours of footage from 198 videos, encompassing a variety of road types and scales, and providing multiple viewing angles for vehicles and pedestrians. The dataset includes over 3.2 million frames, each annotated with weather conditions and scenarios, alongside detailed object annotations derived from extracted images. Capturing complex, high-density scenes of vehicles and pedestrians in crowded urban settings, TSBOW features approximately 71.1 million bounding boxes across eight distinct traffic participant classes. As a robust resource for traffic surveillance research, TSBOW offers substantial potential to deepen insights into traffic dynamics and support advancements in intelligent transportation systems. The initial version focuses on daytime traffic flow under varying weather conditions. Future versions will include ground truth annotations for nighttime scenarios and additional computer vision tasks, such as multi-object tracking, semantic segmentation, vehicle counting, and speed estimation, to further enhance its utility.

	\section*{Acknowledgments}
	This work was supported by the Institute of Information \& communications Technology Planning \& Evaluation (IITP) grant funded by the Korea government (MSIT) (No. 2021-0-01364, An intelligent system for 24/7 real-time traffic surveillance on edge devices).

	\bibliographystyle{unsrtnat}
	\bibliography{references}  

	
	\clearpage
	\setcounter{page}{1}
	\setcounter{equation}{0}
	\setcounter{figure}{0}
	\setcounter{table}{0}
	\setcounter{section}{0}

	\section*{Supplementary Materials}
	
	\begin{figure}[t]
		\centering
		\includegraphics[width=0.8\columnwidth]{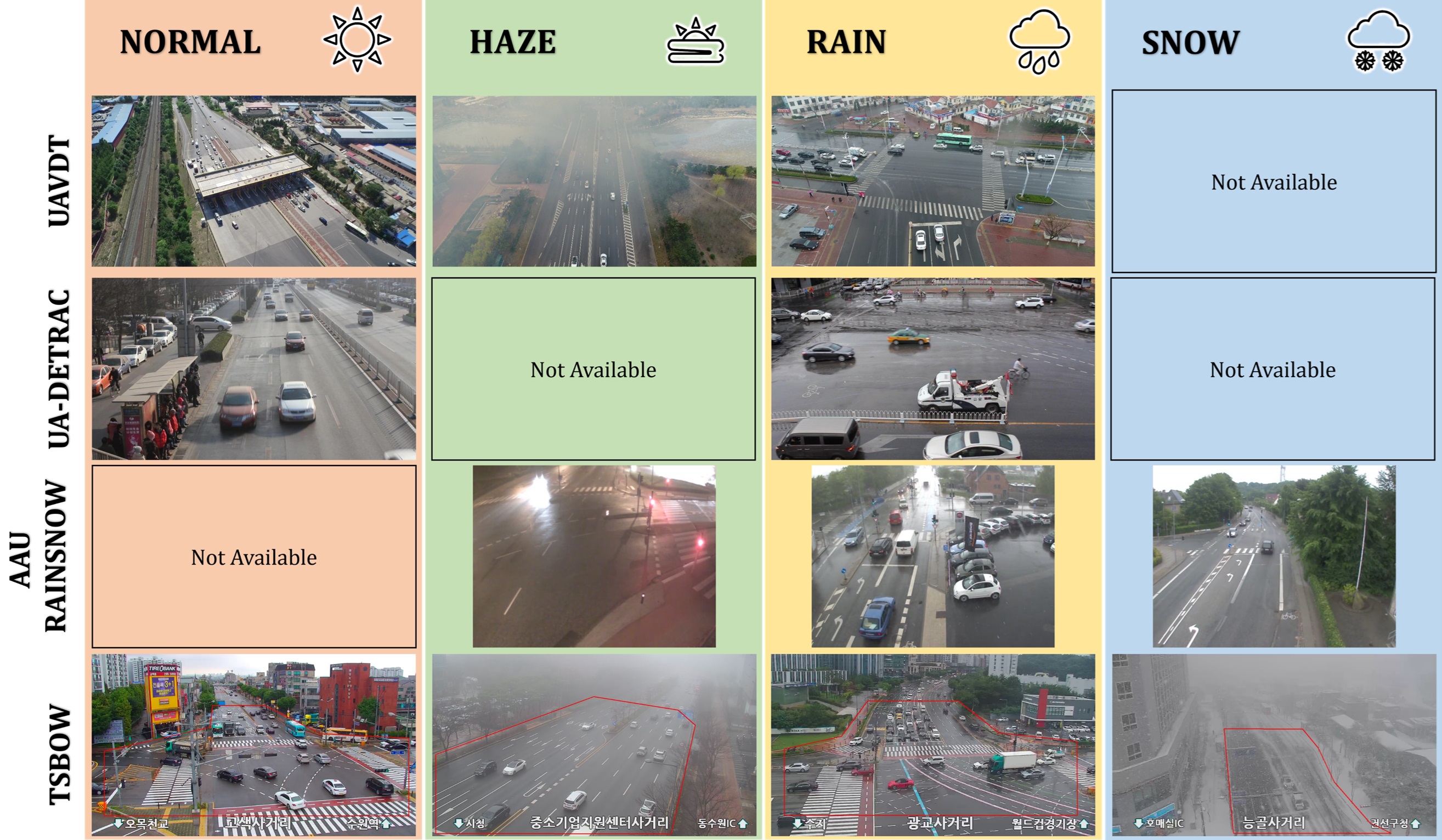}
		\caption{Comparison with other datasets about weather conditions and scales}
		\label{fig:supp_comparison_datasets}
	\end{figure}

	\section{Related Works}
	
	Fig.~\ref{fig:supp_comparison_datasets} compares weather conditions and scales across the TSBOW dataset and other datasets, including UAVDT, UA-DETRAC, and AAURainSnow. UA-DETRAC and AAURainSnow consists solely of fine- and medium-scale videos captured using a color camera, whereas UAVDT, utilizing drone footage, concentrates exclusively on medium and coarse scales. In contrast, the TSBOW dataset encompasses all three scales—fine, medium, and coarse—offering greater diversity.

	\section{The TSBOW Dataset}
	
	\subsection{Labeling Criteria}
	
	Labeling criteria~\cite{label_requirements, label_specifications}, are as follows:
	\begin{itemize}
		\item 
		Bounding boxes must tightly encase objects to accurately capture their shapes and locations, minimizing extraneous background inclusion.
		\item 
		Occluded objects are labeled as fully visible to enable recognition despite partial obscurement.
		\item 
		A vehicle is deemed within the Region of Interest (ROI) if its bounding box center resides therein.
		\item 
		Traffic signs and lights overlapping vehicles are classified as "others" to differentiate background elements from vehicle features.
		\item 
		Objects transported by trucks are separately annotated to ensure detection of conveyed vehicles.
		\item 
		Objects moved by pedestrians are labeled as "others," reflecting their road space occupancy while distinguishing them from standard vehicle categories and aiding differentiation from background elements.
	\end{itemize}
	
	\noindent
	\subsection{Data Format and Description}
	
	The TSBOW dataset comprises videos in MP4 format and extracted frames in JPEG format to minimize re-encoding loss, accompanied by corresponding YOLO label text files for each image. Each label file entry adheres to the bounding box format, specifying the class ID, X and Y coordinates (center point), width, and height. Data for training, validation, and test sets are supplied in text format, supplemented by a YAML file detailing dataset metadata.
	
	\noindent
	\subsection{Train-Test Split}
	
	Videos capture intervals between red-light phases, each lasts approximately two minutes. The three-minute training splits include vehicle flows and queues. The extended five-minute testing splits capture more flows for comprehensive evaluation. Also, varied CCTVs provide diverse viewpoints, mitigating over-optimistic results from similarities.

	\begin{figure}[t]
		\centering
		\includegraphics[width=0.65\columnwidth]{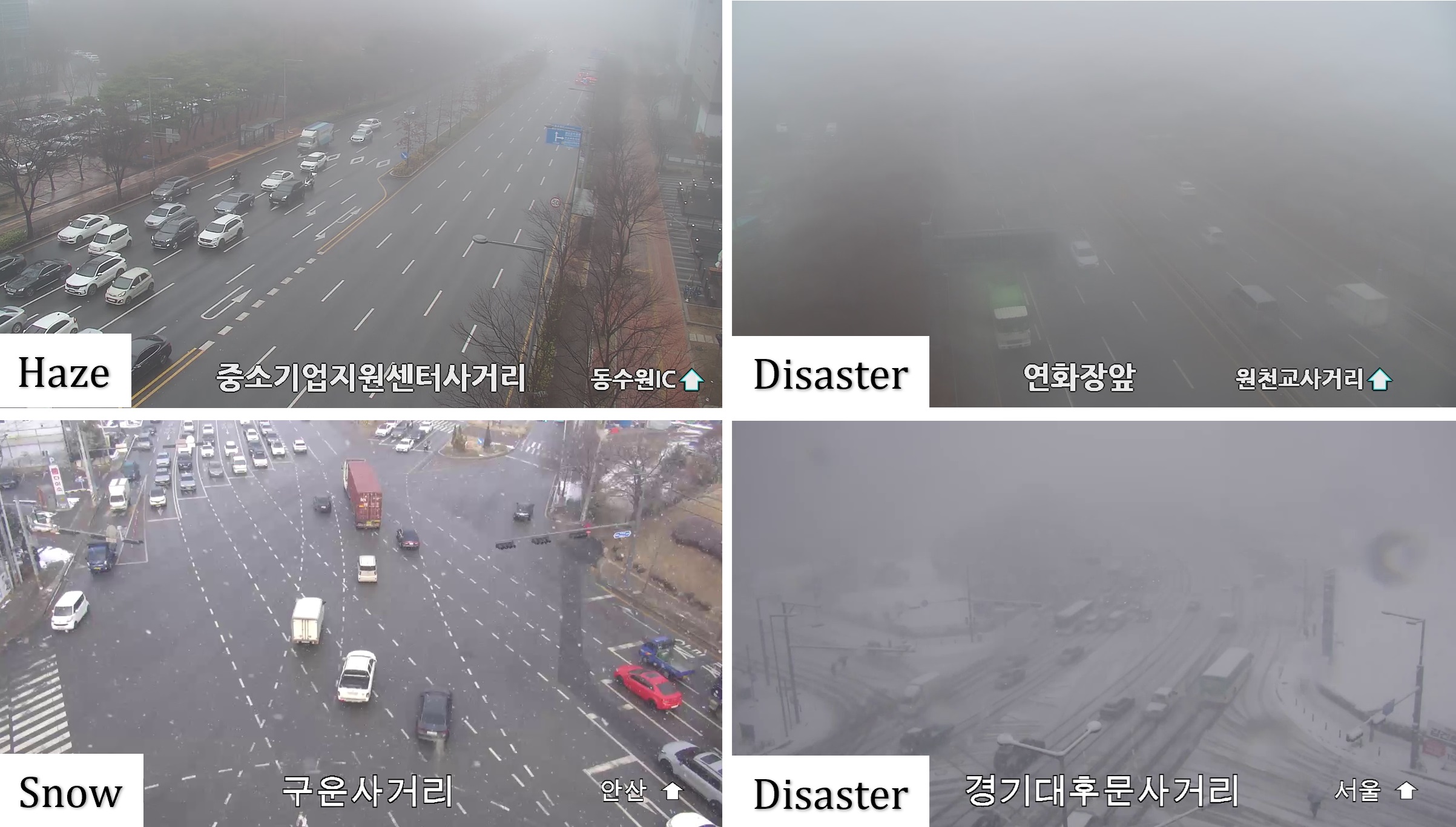}
		\caption{Comparison weather conditions and disaster}
		\label{fig:supp_weather_disaster}
	\end{figure}
	
	
	\begin{figure}[t]
		\centering
		\includegraphics[width=0.65\columnwidth]{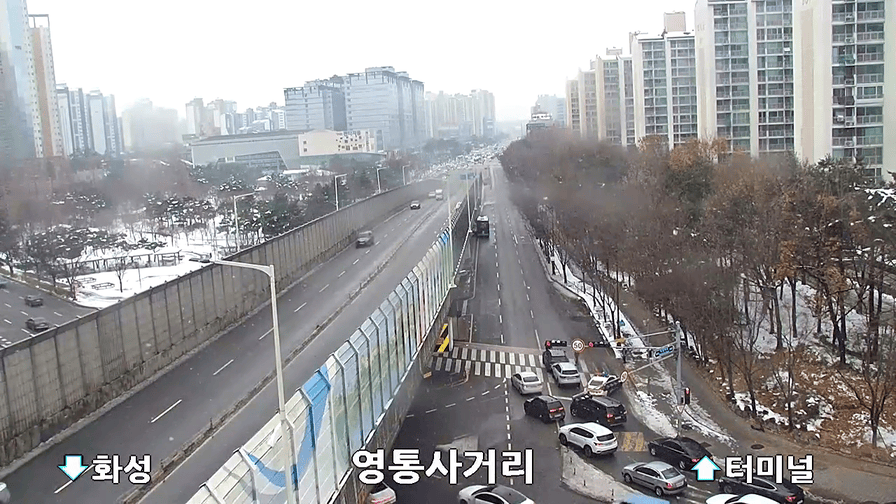}
		\caption{Car accident in the TSBOW dataset}
		\label{fig:supp_accident}
	\end{figure}
	
	\noindent
	\subsection{Differentiation of Weather Conditions and \qq{Disaster} Scenarios}
	
	Fig.~\ref{fig:supp_weather_disaster} distinguishes between \qq{haze} and \qq{snow} in other weather condition and \qq{haze} and \qq{snow} in \qq{disaster} conditions. The fog and snowflakes in \qq{haze} and \qq{snow} have minimal impact on object features and traffic flow. In \qq{disaster} conditions, heavy haze and heavy snow obscures object features, reduces detection accuracy, disrupts traffic, and increases accident frequency.
	
	Additionally, a recorded car accident in snowy conditions highlights the impact of slippery roads (as shown in~\cref{fig:supp_accident}).
	
	\noindent
	\subsection{Detailed Description of Object Classes}
	
	\begin{itemize}
		\item   \textit{car} category includes sedan, SUV, and van, while \textit{bus} includes both small bus and standard bus. 
		\item   \textit{small truck}, comparable in size to SUV or van, comprises pickup truck, small truck, and small box truck. 
		\item   \textit{micromobility} denotes bicycle, motorbike, and scooter, and \textit{pedestrian} annotations are restricted to individuals on crosswalk.
		\item   \textit{truck} is the most diverse vehicle category~\cite{truck_types}, including large box truck, trailer truck, flatbed truck, dump truck, tanker truck, concrete truck, garbage truck, crane truck, tow truck, fire truck, and car transporter.
		\item   In hostile weather conditions, unstable network connections may result in indistinct or corrupted vehicle visuals, which are designated as \textit{unidentified}. 
		\item   Traffic signs and lights partially obscuring vehicles are separately annotated as \textit{others} to distinguish vehicle features from background elements. 
		\item   Vehicles clearly visible yet not fitting predefined categories are also labeled as \textit{others}. 
	\end{itemize}
	
	
	\begin{figure}[t]
		\centering
		\includegraphics[width=0.7\columnwidth]{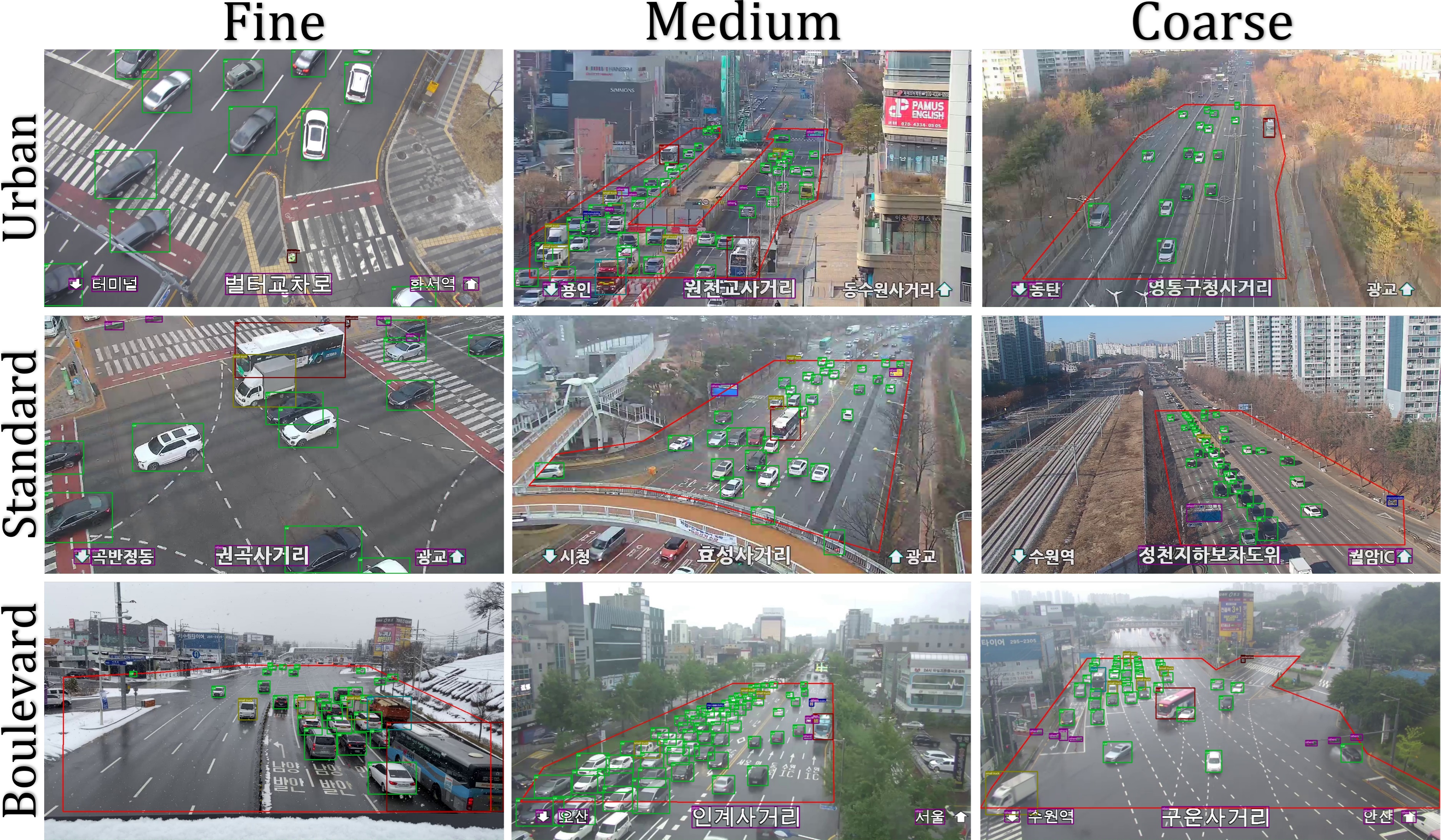}
		\caption{Examples of Road Types and Scales in the TSBOW dataset}
		\label{fig:supp_roadtype_scale}
	\end{figure}
	
	\begin{figure}[t]
		\centering
		\includegraphics[width=\columnwidth]{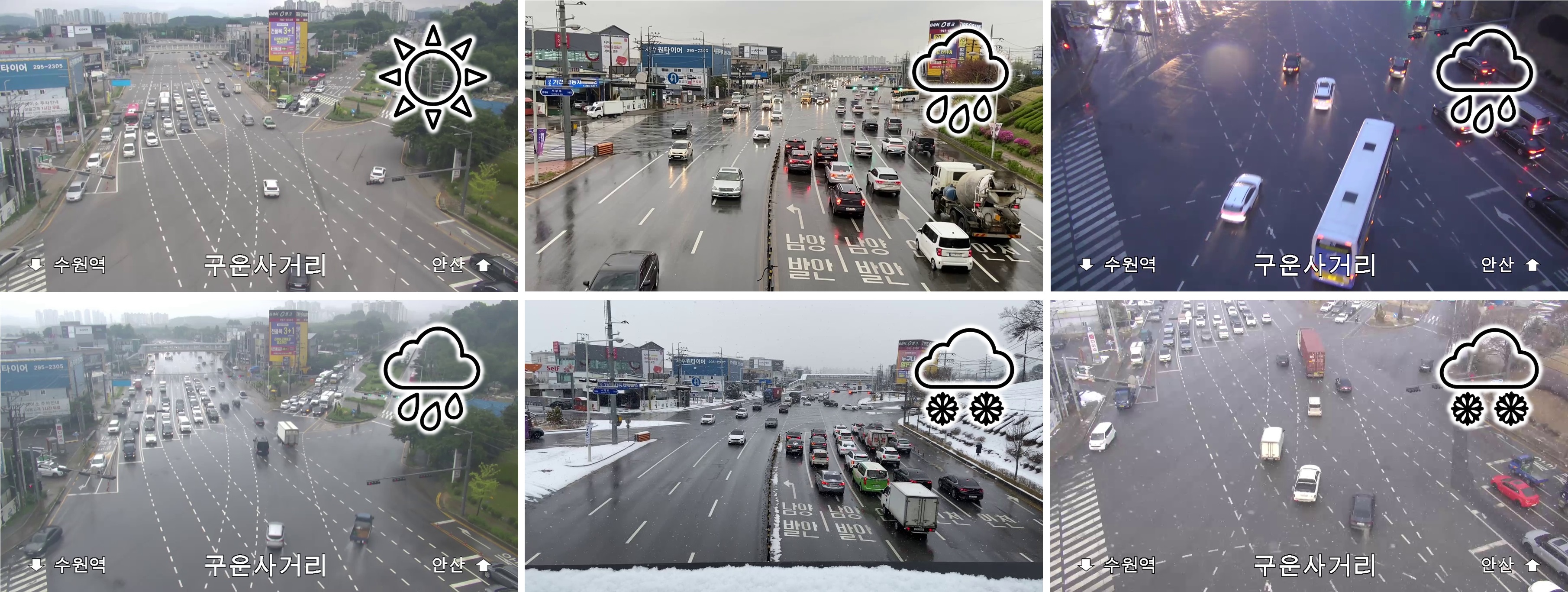}
		\caption{Intersections under different viewpoints in diverse weather conditions}
		\label{fig:supp_fov}
	\end{figure}
	
	\begin{table*}[t]
		\centering
		\setlength{\tabcolsep}{11mm}
		\begin{tabular}{c|c|c|c}
			\hline
			\textbf{Subset} & \textbf{no\_occlusion} & \textbf{light\_occlusion} & \textbf{heavy\_occlusion} \\
			\hline
			\textbf{Train} & 166,437 & 62,090 & 33,996 \\
			\textbf{Val}   & 114,680 & 43,727 & 22,057 \\
			\textbf{Test}  & 440,567 & 160,603 & 86,998 \\
			\hline
			\textbf{Total}   & \textbf{721,684} & \textbf{266,420} & \textbf{143,051} \\
			\hline
		\end{tabular}
		\caption{Distribution of instances by occlusion level across training, validation, and test sets.}
		\label{tab:occlusion_distribution}
	\end{table*}

	\noindent
	\subsection{Diverse Road Types and Scales}
	
	Fig.~\ref{fig:supp_roadtype_scale} illustrates examples of various road types and scales within the TSBOW dataset.
	
	\noindent
	\subsection{Varied Field of View (FOV)}
	
	Fig.~\ref{fig:supp_fov} depicts identical intersections captured from different viewpoints and under diverse weather conditions. The these videos encompass all three scales—fine, medium, and coarse—enhancing its representational diversity.
	
	\noindent
	\subsection{Occlusion Distribution}
	
	Tab.~\ref{tab:occlusion_distribution} details the occlusion distribution of instances across training, validation, and test sets.

	\begin{table*}[t]
		\centering
		\setlength{\tabcolsep}{6.6mm}
		\begin{tabular}{c|c|c|c|c|c}
			\hline
			\textbf{Method} & \textbf{Image Size} & \textbf{Precision} & \textbf{Recall} & \textbf{mAP50} & \textbf{mAP50–95} \\
			\hline
			\multirow{5}{*}{\textbf{YOLOv8x}} 
			& 960  & 0.796 & 0.683 & 0.730 & 0.606 \\
			& 1120 & 0.804 & 0.694 & 0.739 & 0.616 \\
			& 1280 & 0.783 & 0.705 & 0.733 & 0.609 \\
			& 1440 & 0.800 & 0.704 & 0.744 & \textbf{0.620} \\
			& 1600 & 0.798 & 0.705 & \textbf{0.745} & 0.618 \\
			\hline
			\multirow{5}{*}{\textbf{YOLO11x}} 
			& 960  & 0.789 & 0.677 & 0.729 & 0.608 \\
			& 1120 & 0.798 & 0.686 & 0.739 & 0.618 \\
			& 1280 & 0.786 & 0.696 & 0.734 & 0.614 \\
			& 1440 & 0.797 & 0.694 & 0.742 & \textbf{0.620} \\
			& 1600 & 0.797 & 0.698 & 0.743 & \textbf{0.620} \\
			\hline
			\multirow{5}{*}{\textbf{YOLOv12x}} 
			& 960  & \textbf{0.825} & 0.629 & 0.732 & 0.601 \\
			& 1120 & 0.815 & 0.650 & 0.740 & 0.612 \\
			& 1280 & 0.806 & 0.662 & 0.744 & 0.615 \\
			& 1440 & 0.802 & 0.665 & 0.744 & 0.615 \\
			& 1600 & 0.797 & 0.668 & 0.743 & 0.613 \\
			\hline
			\multirow{5}{*}{\textbf{RT-DETR-x}} 
			& 960  & 0.722 & 0.703 & 0.697 & 0.437 \\
			& 1120 & 0.738 & 0.725 & 0.718 & 0.516 \\
			& 1280 & 0.731 & 0.740 & 0.718 & 0.552 \\
			& 1440 & 0.742 & \textbf{0.742} & 0.725 & 0.532 \\
			& 1600 & 0.734 & 0.733 & 0.716 & 0.493 \\
			\hline
		\end{tabular}
		\caption{Validation of model performance with different image sizes on a manually labeled test set.}
		\label{tab:model_performance_image_size}
	\end{table*}
	
	\begin{figure*}[t]
		\centering
		\includegraphics[width=\textwidth]{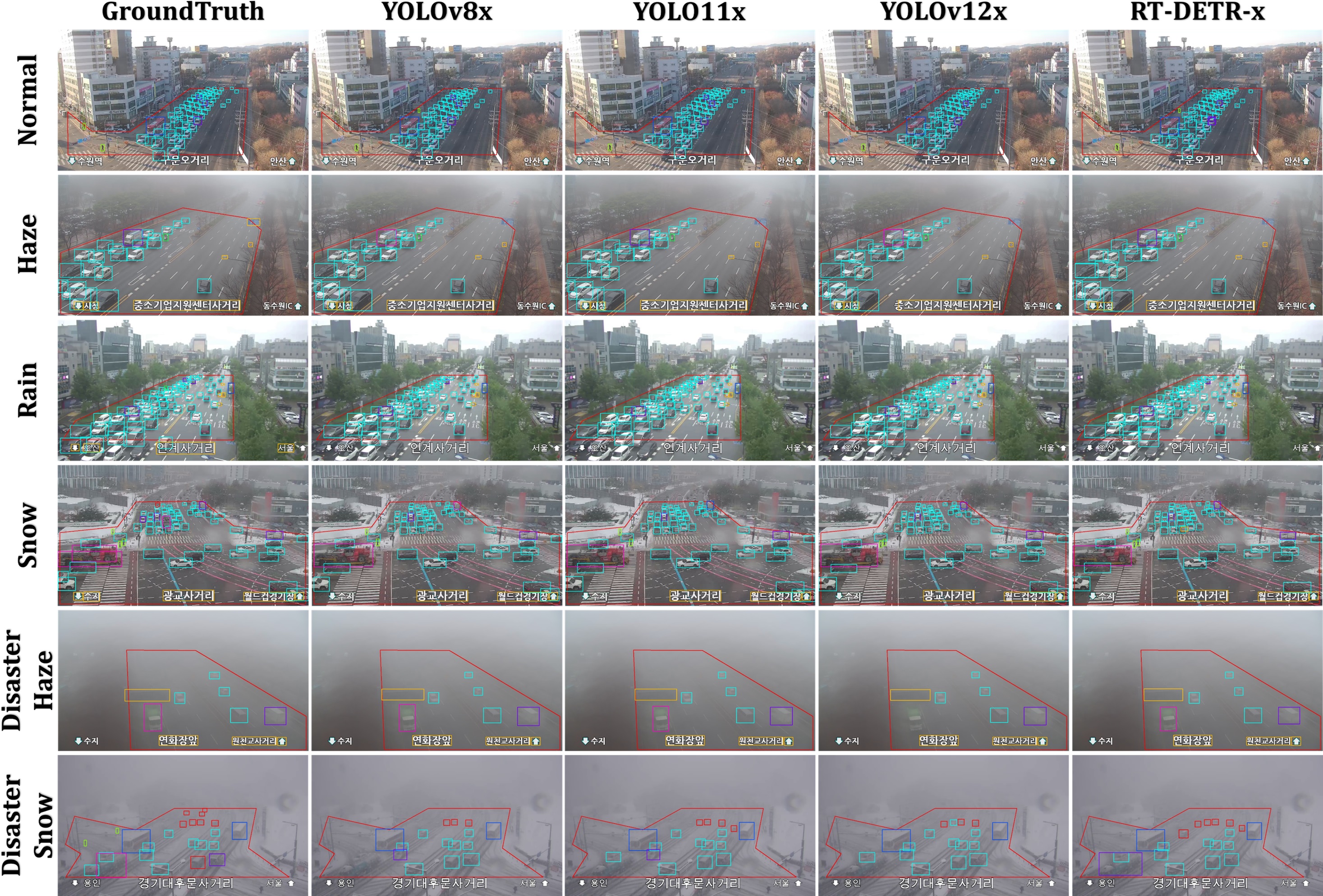}
		\caption{Model performances under different weather conditions}
		\label{fig:supp_models_performance}
	\end{figure*}
	
	\begin{table*}[t]
		\centering
		\setlength{\tabcolsep}{9.5mm}
		\begin{tabular}{c|c|c|c|c}
			\hline
			\textbf{Weather} & \textbf{Scenario} & \textbf{Scale} & \textbf{RoadType} & \textbf{Location} \\
			\hline
			normal & special case & medium & urban & Seoul \\
			haze   & intersection  & medium & urban & Seoul \\
			rain   & intersection  & medium & standard & Seoul \\
			snow   & road          & medium & standard & Suwon \\
			\hline
		\end{tabular}
		\caption{Environmental and road scenario metadata for selected conditions.}
		\label{tab:weather_scenario_metadata}
	\end{table*}
	
	\begin{figure*}[t]
		\centering
		\includegraphics[width=\textwidth]{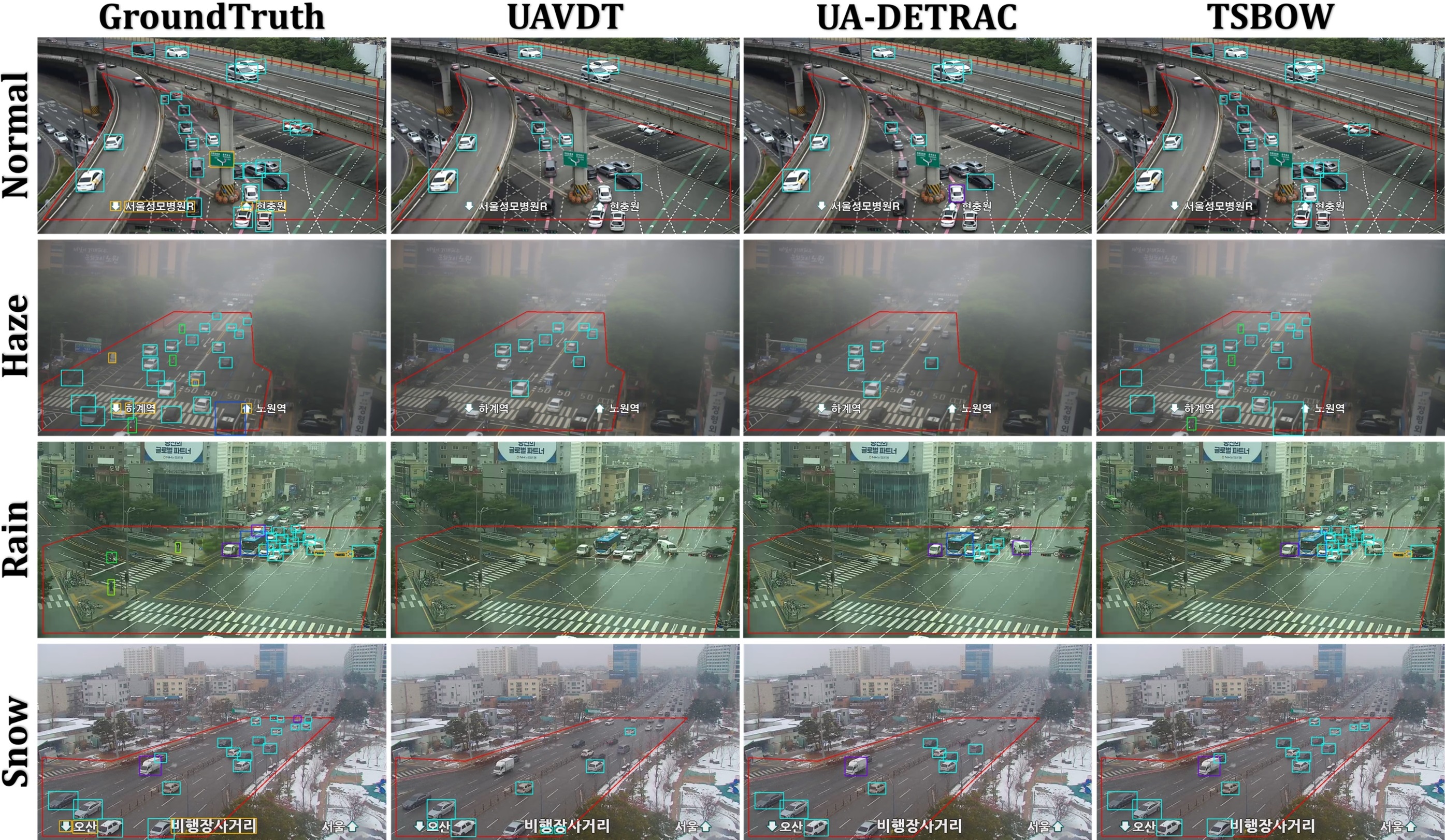}
		\caption{Model performances when training on different datasets}
		\label{fig:supp_comparison_datasets_performance}
	\end{figure*}

	\section{Experiments}
	
	\noindent
	\subsection{Experiment Environment}
	
	Experiments were conducted on a system operating Ubuntu 22.04, equipped with an Intel Core i9-10980XE CPU (3.00GHz), 256GB of RAM, and four NVIDIA RTX A6000 GPUs (each with 48GB of memory). 
	
	\noindent
	\subsection{Model Training and Inference Details}
	\begin{itemize}
		\item 
		For model training, the hyperparameters were: $imgsz=1280, epochs=100, SGD\ optimizer, lr=0.01,$ $ patience=50, momentum=0.937$. For validation, the parameters were: $conf=0.5, IoU=0.6,$ $ max\_detect=300, batch=32$. Validation was conducted across multiple image sizes: 960, 1120, 1280, 1440, and 1680. Tab.~\ref{tab:model_performance_image_size} provides detailed model performance after validating with different image sizes.
		
		\item
		For inference on the remaining frames, $conf=0.6$ was applied to small objects (pedestrian and micromobility), while $conf=0.3$ was used for other classes. The higher confidence score for pedestrians addresses challenges associated with significant overlap when crossing streets, whereas the lower score for other classes enhances the detection of small objects, particularly those near the boundaries of the region of interest (ROI).
	\end{itemize}
	
	\noindent
	\subsection{Model Performances under Different Weather Conditions}
	
	Fig.~\ref{fig:supp_models_performance} presents the performance of models on a manually labeled test set under various weather conditions, following training for 100 epochs on the training set.
	
	\noindent
	\subsection{Impact of \qq{Disaster} Scenarios on Model Performance}
	
	In extreme weather conditions, vehicles often appear as indistinct shadows, leading object detection models to misclassify them as part of the background.
	\begin{itemize}
		\item 
		Small vehicles at a distance during heavy snowfall frequently appear as gray polygons devoid of distinct features. Consequently, models often fail to classify these as objects, even when an "unidentified" class is available. Additionally, white vehicles are frequently undetected in snowy conditions due to their color blending with the snow-covered background. The pre-trained models exhibit poor performance in such conditions, whereas \cref{fig:supp_models_performance} illustrates improved detection accuracy after fine-tuning. Pre-trained models commonly overlook "unidentified" objects, leading users to erroneously perceive the road as empty or sparsely traffic, despite potentially high vehicle volumes on the route. This misdetection has significant implications, as it impedes the government's ability to promptly manage traffic in response to accidents or congestion caused by degraded road surface conditions.
		
		\item 
		In conditions of extreme haze, image quality is severely compromised by heavy noise, obscuring vehicle features. As a result, models typically detect only those vehicles in close proximity to the camera. Vehicles at greater distances are often missed, as only their lights remain visible, while other parts blend into the hazy background.
	\end{itemize}
	
	\noindent
	\subsection{Model Performance Across Datasets}
	
	Tab.~\ref{tab:weather_scenario_metadata} details the comparison set, encompassing weather conditions, scenarios, scales, road types, and locations. As the "medium" scale is common across all datasets, it is selected for the comparison set.
	
	Fig.~\ref{fig:supp_comparison_datasets_performance} illustrates a comparative analysis of model performance when trained on different datasets.

\end{document}